%% file: main.tex
\pdfoutput=1

\documentclass[11pt]{article}

\usepackage[final]{acl}
\usepackage{todonotes}
\usepackage{times}
\usepackage{latexsym}
\usepackage{amsmath}

\usepackage[T1]{fontenc}

\usepackage[utf8]{inputenc}

\usepackage{microtype}

\usepackage{inconsolata}

\usepackage{graphicx}
\usepackage{multirow}
\usepackage{contour}

\input{macros}
\usepackage{arabtex}

\usepackage{utf8}
\usepackage{epstopdf}
\usepackage{pdfpages}

\usepackage{alltt}
\newcommand{\AR}[1]{\<#1>}
\usepackage{enumitem}
\usepackage{multicol}
\usepackage{needspace}
\usepackage{placeins}
\usepackage{tcolorbox}
\tcbuselibrary{breakable,skins}
\newtcolorbox{promptpart}[1]{
  enhanced,
  colback=black!1,
  colframe=blue!45!black,
  colbacktitle=blue!7,
  coltitle=black,
  title={#1},
  fonttitle=\bfseries,
  boxrule=0.6pt,
  arc=1mm,
  left=2.5mm,
  right=2.5mm,
  top=1.5mm,
  bottom=1.5mm,
  before={\par\addvspace{7pt}\noindent\nolinenumbers},
  after={\par\linenumbers\addvspace{7pt}}
}
\usepackage{booktabs} 

\title{Arabic Morphosyntactic Tagging and Dependency Parsing\\ with Large Language Models}

\author{Mohamed Adel\textsuperscript{1} \quad Bashar Alhafni\textsuperscript{2} \quad Nizar Habash\textsuperscript{1} \\
        Computational Approaches to Modeling Language Lab \\
        \textsuperscript{1}New York University Abu Dhabi \\
        \textsuperscript{2}Mohamed bin Zayed University of Artificial Intelligence \\
  \texttt{\{mohamed.adel,nizar.habash\}@nyu.edu, bashar.alhafni@mbzuai.ac.ae}
  }

\begin{document}
\maketitle

\setcode{utf8}
\vocalize




\begin{abstract}
LLMs perform strongly across NLP, but their ability to produce explicit grammatical analyses remains unclear. Arabic provides a challenging testbed due to its rich morphology and orthographic ambiguity, which create strong morphology–syntax interactions. We present a unified evaluation of LLMs on Arabic morphosyntactic tagging and dependency parsing, covering pre-tokenized, raw-text, and cascaded settings. We compare zero-shot prompting with retrieval-based in-context learning. Relevant demonstrations substantially improve performance. The strongest LLMs approach supervised tagging and parsing systems; however, they require substantial annotated data for demonstration retrieval and considerable computational resources. We make all code and data used in this paper publicly available.\footnote{\url{https://github.com/CAMeL-Lab/arabic-morphosyntax-llm}}

\end{list}
\end{abstract}

\section{Introduction}

 LLMs achieve strong performance across many NLP tasks  \citep{openai2024gpt4technicalreport,geminiteam2025geminifamilyhighlycapable, grattafiori2024llama3herdmodels, yang2025qwen3technicalreport}. However, it remains unclear whether they can reliably produce structured linguistic analyses such as morphosyntactic features and dependency trees. Prior work shows that LLMs can recover some linguistic patterns when appropriately prompted, but performance remains uneven when tasks require explicit morphosyntactic constraints or precise structured predictions such as parsing \citep{blevins-etal-2023-prompting,anh-etal-2024-morphology,miaschi-etal-2024-evaluating,tian-etal-2024-large,ezquerro-etal-2025-better}.

Morphologically rich languages provide a particularly challenging testbed for evaluating the structural competence of LLMs \citep{anh-etal-2024-morphology, ismayilzada-etal-2025-evaluating}. Although recent work has increasingly examined the linguistic capabilities of LLMs, morphosyntactic tagging and dependency parsing are often evaluated separately, despite evidence that morphological and POS information substantially affect parsing and that joint evaluation can reveal weaknesses hidden in isolated or gold-preprocessed settings \citep{tsarfaty-etal-2012-joint,zhang-etal-2015-randomized,10.1007/978-3-030-32381-3_50, more-etal-2019-joint,temesgen-etal-2025-extracting, matsuda-etal-2025-step}.

In this paper, we evaluate whether instruction-tuned LLMs can perform morphosyntactic and syntactic analysis for Standard Arabic, a language whose rich inflectional morphology, cliticization, and tokenization ambiguities create strong morphology-syntax interactions \citep{habash-rambow-2005-arabic, green-manning-2010-better, marton-etal-2013-dependency}.
%
We evaluate instruction-tuned LLMs on two tasks: \textbf{(i) morphosyntactic tagging} with 14 morphological features and \textbf{(ii) labeled dependency parsing}, comparing zero-shot prompting with retrieval-based in-context learning (ICL) \citep{brown2020languagemodelsfewshotlearners, liu-etal-2022-makes, min-etal-2022-rethinking}. We also examine whether model-predicted morphology can improve dependency parsing. 
Results show that prompt design and example retrieval strongly affect performance:
open-weight models struggle in zero-shot but improve with retrieved demonstrations, while proprietary LLMs approach specialized supervised taggers and parsers.

Our contributions are as follows:
\begin{itemize}[noitemsep,topsep=0pt,parsep=0pt,partopsep=0pt]
\item A unified evaluation of morphosyntactic tagging and dependency parsing across pre-tokenized, raw-text, and cascaded settings.
\item An analysis of prompt design and retrieval-based ICL for structured linguistic prediction.
\item A detailed error analysis identifying which aspects of Arabic tagging and parsing LLMs capture reliably and which remain difficult.
\item A SOTA extension of CamelParser, a supervised Arabic dependency parser \citep{elshabrawy-etal-2023-camelparser2}.\footnote{\url{https://github.com/CAMeL-Lab/camel_parser}}
\end{itemize}

\section{Related Work}
\label{sec:related}

\subsection{Arabic Tagging and Parsing}

\paragraph{Arabic Morphosyntactic Tagging} 
Arabic morphological modeling has proven useful for various downstream tasks, including machine translation \cite{sadat-habash-2006-combination,el-kholy-habash-2010-orthographic}, grammatical error correction \cite{alhafni-etal-2023-advancements}, and text rewriting \cite{alhafni-etal-2022-user}. A wide range of approaches have been proposed for Arabic morphosyntactic tagging \cite{diab-etal-2004-automatic,habash-rambow-2005-arabic,pasha-etal-2014-madamira,abdelali-etal-2016-farasa,zalmout-habash-2017-dont,inoue-etal-2017-joint,zalmout-etal-2018-noise,zalmout-habash-2019-adversarial,zalmout-habash-2020-joint,khalifa-etal-2020-morphological}. More recently, \citet{inoue-etal-2022-morphosyntactic} introduced BERT-based models that achieve state-of-the-art (SOTA) performance on MSA and multiple Arabic dialects, and were later integrated into CAMeL Tools \cite{obeid-etal-2020-camel}. Much of this work relies on the Penn Arabic Treebank (PATB), which serves as a central benchmark for Arabic morphological and syntactic annotation \citep{Maamouri:2004:penn}. To the best of our knowledge, the use of LLMs for Arabic morphosyntactic tagging has not been systematically investigated. In this work, we evaluate LLMs on PATB and compare them against CAMeL Tools as a strong benchmark system.


\paragraph{Arabic Dependency Parsing} Arabic dependency parsing has benefited from the development of several treebanks \cite{Smrvz:2002:prague,Maamouri:2004:penn,habash-roth-2009-catib,Dukes:2010:dependency,taji-etal-2017-universal,nivre-etal-2020-universal,al-ghamdi-etal-2021-dependency}. More recently, \citet{habash-etal-2022-camel} introduced the CAMeL Treebank (CAMeLTB), a multi-genre Arabic dependency treebank created according to the Columbia Arabic Treebank (CATiB) formalism \cite{habash-roth-2009-catib}. A wide range of parsing approaches have been proposed for Arabic \cite{green-manning-2010-better,marton-etal-2013-dependency,shahrour-etal-2016-camelparser,straka-2018-udpipe,kankanampati-etal-2020-multitask,ghamdi:2023:fine,elshabrawy-etal-2023-camelparser2}. Among these, \citet{elshabrawy-etal-2023-camelparser2} introduced CAMeLParser2.0, a BERT-based parser that currently achieves state-of-the-art performance on CAMeLTB. In this work, we benchmark LLMs on CAMeLTB and compare their performance against CAMeLParser2.0.

Recently, \citet{alsabahi-et-al-2026-i3rabLLM} explored the use of LLMs for Arabic syntactic analysis.
They evaluate LLMs on a carefully curated dataset targeting specific grammatical phenomena and frame the task as morphosyntactic annotation over gold-tokenized text. Our work differs in several ways. First, while their task predicts a morphosyntactic label for each token, we use a 14-dimensional morphosyntactic representation and extend the task to full dependency parsing with labeled relations. Second, their dataset assumes gold tokenization, whereas we evaluate parsing with both gold and model-predicted tokenization. Finally, their dataset targets specific constructions, while our experiments use multi-genre treebank data reflecting naturally occurring Arabic text. We also evaluate a wider range of prompting strategies and compare LLMs against a SOTA parsing system.

\input{Example}


\subsection{LLMs and Linguistic Knowledge}
Recent work has examined whether LLMs encode linguistic knowledge that can be explicitly elicited through prompting \cite{blevins-etal-2023-prompting,miaschi-etal-2024-evaluating}. A parallel line of work evaluates LLMs on morphological generalization, showing mixed results, particularly for morphologically rich languages \cite{weissweiler-etal-2023-counting,anh-etal-2024-morphology,weller-di-marco-fraser-2024-analyzing,Waldis:2024:holmes,ismayilzada-etal-2025-evaluating,temesgen-etal-2025-extracting}. Other studies investigate syntactic structure and parsing, finding that while LLMs capture aspects of shallow syntax, full dependency parsing remains challenging \cite{tian-etal-2024-large,ezquerro-etal-2025-better,zhang-etal-2025-self-correction}.

ICL enables LLMs to perform new tasks from a small set of demonstrations without parameter updates \citep{brown2020languagemodelsfewshotlearners}. Prior work shows that its effectiveness depends strongly on the choice and ordering of examples, with retrieval-based selection often outperforming random sampling and demonstrations helping models infer both the task and the expected output format \citep{liu-etal-2022-makes,min-etal-2022-rethinking,rubin-etal-2022-learning}. Motivated by this line of work, we evaluate whether instruction-tuned LLMs can produce explicit morphosyntactic analyses and labeled dependency trees for Arabic under both zero-shot and in-context settings, and how different example-selection strategies affect performance. In particular, we study how prompting and example selection affect the ability of LLMs to generate structured linguistic outputs for Arabic morphosyntactic tagging and dependency parsing.

\section{Tasks}
\label{sec:tasks}



\subsection{Task 1: Morphosyntactic Tagging}
\label{sec:tag_task}

\paragraph{Task Definition}
In morphosyntactic tagging, the model receives an Arabic sentence as a fixed sequence of tokens and outputs a complete morphosyntactic analysis for each token. Specifically, it predicts values for 14 categorical features: part-of-speech (\texttt{pos}), aspect (\texttt{asp}), mood (\texttt{mod}), voice (\texttt{vox}), person (\texttt{per}), \textcolor{black}{form gender (\texttt{gen}), form number (\texttt{num})}, case (\texttt{cas}),  state (\texttt{stt}), proclitics (\texttt{prc3}, \texttt{prc2}, \texttt{prc1}, \texttt{prc0}) and the pronominal enclitic (\texttt{enc0}). See example in Table~\ref{tab:morphtag}.
This feature inventory follows common Arabic morphological disambiguation practice  \citep{pasha-etal-2014-madamira, obeid-etal-2020-camel}.

\paragraph{Prompt Formulation}
The prompt (Appendix~\ref{app:mtag_prompt}) provides the sentence as a fixed list of whitespace/punctuation tokens and supplies the target feature inventory, with brief descriptions and allowed values. The model returns a complete morphosyntactic analysis for each token without changing the given list or its order.

\subsection{Task 2: Dependency Parsing}
\label{sec:parse_task}
\paragraph{Task Definition}

In dependency parsing, the model predicts a labeled dependency tree over a sentence's tokens. For each token, it identifies the syntactic head, using index 0 for the root, and assigns a dependency relation label. We adopt the CATiB representation \citep{habash-roth-2009-catib,habash-etal-2022-camel}. CATiB follows the ATB tokenization standard, which separates all clitics except the definite article (Figure~\ref{fig:parse_example}). We evaluate parsing under two input conditions: gold tokenization and raw text (end-to-end).

\paragraph{Prompting Formulation}
The parsing prompts (Appendix~\ref{app:parsing_prompts}) provide the input sentence, the CATiB dependency-label inventory, and annotation guidelines. The model either preserves the supplied gold sequence of morphological tokens or derives the morphological tokenization from raw text according to CATiB conventions before producing the dependency analysis.

\subsection{Task Integration}
\label{sec:task_integration}
We connect the two tasks on the same PATB sentences and their converted CATiB dependency trees. In the independent condition, morphology and raw-text parsing are predicted separately; in the cascaded condition, the same model's morphology predictions for the input tokens are also provided to the parser as auxiliary evidence. The cascaded prompt reuses the two task prompts (Appendix~\ref{app:cascaded_prompt}).
We also evaluate a unified NUDAR/UD formulation for UPOS, morphology, and dependencies (Appendix~\ref{app:nudar_ud}), but center PATB/CATiB because they are linguistically motivated for Arabic, supported by manually annotated treebanks, and widely used in Arabic NLP.

\section{Methodology}
\label{sec:methodology}

We evaluate LLMs on two structured tasks: \textbf{morphosyntactic tagging} and \textbf{dependency parsing} (\S\ref{sec:tasks}), comparing zero-shot prompting with ICL and varying example selection strategies.

\paragraph{In-Context Learning Setup}
We evaluate ICL by varying both the number of examples (shots) and the strategy used to select them. For each test instance $x$ (from the dev or test splits), we retrieve examples from a fixed candidate pool $\mathcal{D}_{\text{train}}=\{(x_i,y_i)\}$ consisting only of training data (\S\ref{sec:setup}). Selection relies only on the input side $x_i$ (raw text or token sequence, depending on the task and setting), ensuring that no dev or test labels influence retrieval. Formally, let $k$ denote the number of examples prepended to $x$. We vary two factors: the \textbf{shot count} $k \in \{0,1,3,5,10\}$, where $k=0$ corresponds to zero-shot prompting, and the \textbf{selection method} $m$, which determines which $k$ examples from $\mathcal{D}_{\text{train}}$ are included in the prompt. Each configuration $(k,m)$ therefore defines a prompt $P(x;k,m)$ for each test instance $x$. 





\paragraph{Selection Methods}
We explore two example-selection methods based on surface- and semantic-level similarity. \textbf{Surface similarity} retrieves inputs that are close at the character or subword level, which can capture shared orthographic patterns and morphological features in Arabic even when exact word overlap is limited. We implement this using chrF++ \cite{popovic-2015-chrf,popovic-2017-chrf}, following \citet{ginn-etal-2024-teach}, who showed that substring-based similarity better reflects morphological relatedness than word-level overlap. \textbf{Semantic similarity} retrieves inputs that are close in meaning even when surface forms differ. We implement this by embedding each input using CAMeLBERT-MSA \cite{inoue-etal-2021-interplay} and ranking candidates by cosine similarity.

After computing similarity scores, we consider two retrieval regimes: \textbf{high-similarity} (near-neighbor) and \textbf{low-similarity} (far-neighbor). 
This allows us to distinguish improvements driven by content similarity from those resulting from general format calibration.

In addition, we evaluate \textbf{random selection}, where $k$ examples are sampled randomly from  $\mathcal{D}_{\text{train}}$. Overall, this yields five example-selection methods: random sampling, and for each similarity measure (surface and semantic) we evaluate both high- and low-similarity retrieval.

\section{Experimental Setup}
\label{sec:setup}

This section describes the experimental setup used to evaluate LLMs on Arabic morphosyntactic tagging and dependency parsing. We present the datasets used in our experiments, the evaluated systems, and the evaluation metrics for each task.

\subsection{Data}
\label{subsec:data}

\paragraph{Morphosyntactic Tagging}
We evaluate morphosyntactic tagging on the PATB (MSA, newswire) \citep{Maamouri:2004:penn} following the morphological tag representations of \newcite{inoue-etal-2022-morphosyntactic}, 
and Train/Dev/Test splits based on \newcite{Diab:2013:ldc}.  
Summary statistics are in Table~\ref{tab:data_stats}.

\paragraph{Dependency Parsing}
We evaluate dependency parsing in two data configurations. For the gold-tokenized and raw-text parsing experiments, we use CAMeLTB \citep{habash-etal-2022-camel}, which is annotated in the CATiB dependency representation \citep{habash-roth-2009-catib}.
We adopt the Train/Dev/Test splits following \citet{habash-etal-2022-camel}. To compare independent and cascaded processing, we use CATiB dependency trees converted from the PATB constituency trees \citep{taji-etal-2017-universal}. These trees correspond to the original PATB Test sentences and are therefore aligned with their PATB morphosyntactic annotations. We refer to this paired evaluation data as PATB--CATiB. Summary statistics are in Table~\ref{tab:data_stats}, and Table~\ref{tab:data_task_coverage} summarizes the task coverage of each data configuration.

\begin{table}[t]
\centering
\small
\setlength{\tabcolsep}{5pt}
\begin{tabular}{lcc}
\toprule
\textbf{Data} & \shortstack{\textbf{Morphosyntactic}\\\textbf{Tagging}} & \shortstack{\textbf{Dependency}\\\textbf{Parsing}} \\
\midrule
PATB        & \textbf{Yes} & -- \\
CAMeLTB     & -- & \textbf{Yes} \\
PATB--CATiB & \textbf{Yes} & \textbf{Yes} \\
\bottomrule
\end{tabular}
\caption{Task coverage of PATB, CAMeLTB, and PATB--CATiB. In PATB--CATiB, morphosyntactic tagging is assessed on PATB words and parsing on CATiB tree tokens converted from the same PATB sentences.}
\label{tab:data_task_coverage}
\end{table}

\input{data_stats}

\subsection{Systems}
\label{subsec:systems}

\paragraph{LLMs}
We benchmark multiple state-of-the-art LLMs with strong multilingual capabilities, including Arabic. Our evaluation includes two open-weight models, {Llama~4~Scout}~\citep{meta2025llama4} (\textbf{Llama4}) and {Qwen3-Next-80B-A3B-Thinking}~\citep{yang2025qwen3technicalreport} (\textbf{Qwen3}), as well as two proprietary models, {Gemini~3~Flash}~\citep{geminiteam2025geminifamilyhighlycapable} (\textbf{Gemini3}) and {GPT-5.2}~\citep{openai2026gpt5systemcard} (\textbf{GPT5.2}). 
Arabic-centric models (ALLaM, Fanar, Jais) were not included because their available context lengths were less suitable for our long-context ICL setup \citep{bari2025allam,fanarteam2025fanararabiccentricmultimodalgenerative,sengupta2023jaisjaischatarabiccentricfoundation}.

\paragraph{Supervised Baselines}
For morphosyntactic tagging, we use \textbf{CAMeL Tools} \citep{obeid-etal-2020-camel}, which was trained on PATB. For dependency parsing, we evaluate the released \textbf{CamelParser2.0}, trained on PATB and CAMeLTB \citep{elshabrawy-etal-2023-camelparser2}, and \textbf{CamelParser2.1}, our decoding extension of the same checkpoint. CamelParser2.0 enforces exactly one root per input, whereas CamelParser2.1 uses projective multiroot decoding while continuing to require at least one root and to prohibit cycles and self-loops.

\subsection{ICL Configuration}
We use the Dev sets to select the best ICL configuration. First, all candidate LLMs are evaluated in the zero-shot setting on the Dev set for each task, alongside the supervised baselines. The strongest open-weight LLM under zero-shot performance is chosen as the ICL exploration model. Using this model, we explore the ICL configuration space by varying both the shot count $k$ and the example-selection method described in \S\ref{sec:methodology}. The best-performing configuration ($k$\textsuperscript{*},$m$\textsuperscript{*}) on the Dev set is fixed and used to benchmark on the test sets.

\subsection{Evaluation Metrics}
\label{subsec:metrics}

\paragraph{Morphosyntactic Tagging}
Our primary tagging metric is \textbf{All Tags}, a strict token-level exact match score that counts a token as correct only if all 14 morphosyntactic features are predicted correctly. We use \textbf{All Tags} as the main score because it best reflects end-to-end correctness of the full feature structure. We also report \textbf{Tag F\textsubscript{1}}, which captures partial correctness by aggregating matches over individual feature values across the 14-feature inventory (macro average).

\paragraph{Dependency Parsing}
Our primary parsing metric is \textbf{LAS} (Labeled Attachment Score), which measures the percentage of tokens whose predicted head and dependency relation label match the gold parse. We additionally report \textbf{UAS} (Unlabeled Attachment Score), which evaluates head correctness irrespective of the label, and \textbf{LS} (Label Score), which evaluates relation label correctness irrespective of the head. In the raw-text parsing condition, we also report \textbf{Tok F\textsubscript{1}} (Tokenization F\textsubscript{1}), which compares the aligned predicted and gold token sequences: matching aligned token forms indicate correct boundaries, while precision and recall penalize extra and missing splits. We use the CAMeL CoNLL evaluator.\footnote{\url{https://github.com/CAMeL-Lab/camel_conll}}

\paragraph{Joint PATB--CATiB Evaluation}
Here, joint means that morphology and parsing are evaluated on the same aligned PATB sentences, not that they are jointly trained or predicted simultaneously. Morphology is scored on the original PATB tokens using \textbf{All Tags} and \textbf{Tag F\textsubscript{1}}, whereas tokenization and parsing are scored on the predicted and gold CATiB token sequences using \textbf{Tok F\textsubscript{1}}, \textbf{UAS}, \textbf{LS}, and \textbf{LAS}. This keeps each task on its appropriate annotation unit.

\section{Results}
\label{sec:results}


We first report the two tasks introduced in Section~\ref{sec:methodology}. For each task, we present Dev results used for model and ICL configuration selection, followed by Test results under zero-shot and best-ICL. We then connect the tasks in the joint PATB--CATiB evaluation.

\begin{table}[t!]
\centering
\small
\tabcolsep8pt
\begin{tabular}{lcc}
\toprule
\textbf{Model} & \textbf{All Tags} & \textbf{Tag F\textsubscript{1}} \\
\midrule
Llama4   & 8.9  & 81.9 \\
Qwen3    & 14.1 & 82.8 \\
GPT5.2   & 39.9 & 89.7 \\
Gemini3  & 48.4 & 92.6 \\
\midrule
CAMeL Tools & \textbf{96.2} & \textbf{99.6} \\
\bottomrule
\end{tabular}
\caption{\textbf{Morphosyntactic Tagging} results on the PATB Dev set (zero shot).}
\label{tab:mtag_dev_zeroshot}
\end{table}


\begin{table*}[t!]
\centering
\small
\tabcolsep3pt
\begin{tabular}{lcccccccccc}
\toprule
 \textbf{Number of Shots} & \multicolumn{2}{c}{\textbf{0}} & \multicolumn{2}{c}{\textbf{1}} & \multicolumn{2}{c}{\textbf{3}} & \multicolumn{2}{c}{\textbf{5}} & \multicolumn{2}{c}{\textbf{10}} \\
\cmidrule(lr){2-3} \cmidrule(lr){4-5} \cmidrule(lr){6-7} \cmidrule(lr){8-9} \cmidrule(lr){10-11}
\textbf{Selection Method} & \textbf{All Tags} & \textbf{Tag F\textsubscript{1}} & \textbf{All Tags} & \textbf{Tag F\textsubscript{1}}  & \textbf{All Tags} & \textbf{Tag F\textsubscript{1}}  & \textbf{All Tags} & \textbf{Tag F\textsubscript{1}}  &\textbf{All Tags} & \textbf{Tag F\textsubscript{1}}  \\
\midrule
Lowest Cosine Similarity & -- & -- & 16.5 & 83.5 & 20.1 & 84.9 & 23.3 & 86.1 & 29.0 & 87.8 \\
Lowest chrF++            & -- & -- & 21.4 & 82.1 & 23.6 & 85.5 & 24.5 & 86.2 & 23.9 & 86.2 \\
Random                   & 14.1 & 82.8 & 45.0 & 90.8 & 56.7 & 93.1 & 59.4 & 92.7 & 65.9 & 95.0 \\
Highest Cosine Similarity& -- & -- & 59.2 & \textbf{93.5} & 68.4 & 95.3 & 72.0 & 96.0 & 76.0 & 96.8 \\
Highest chrF++           & -- & -- & \textbf{59.5} & 92.6 & \textbf{70.6} & \textbf{95.7} & \textbf{74.1} & \textbf{96.4} & \textbf{78.0} & \textbf{97.2} \\
\bottomrule
\end{tabular}
\caption{\textbf{Morphosyntactic Tagging} performance comparison of shot selection methods across different numbers of shots for \textbf{Qwen3} on the PATB Dev set. The $k{=}0$ (zero-shot) results correspond to the \textit{Random} row, since no retrieval is performed in that setting.}
\label{tab:selection_methods}
\end{table*}


\subsection{Morphosyntactic Tagging}
\label{subsec:results_mtag}

\paragraph{Baseline}
Table~\ref{tab:mtag_dev_zeroshot} reports zero-shot dev results. The supervised baseline (CAMeL Tools) is near ceiling and substantially outperforms all LLMs, underscoring the difficulty of predicting complete 14-feature bundles for each token. Among LLMs, Gemini3 performs best, followed by GPT5.2. Although some open-weight models obtain relatively higher feature-level Tag F\textsubscript{1}, their All Tags scores remain much lower, showing how small feature errors compound under exact bundle matching. We therefore select Qwen3, the strongest open-weight model on dev, for ICL analysis.

\paragraph{ICL}
Table~\ref{tab:selection_methods} shows that ICL yields large gains even with very few demonstrations. With random 1-shot prompting, All Tags improves by 30.9 points over zero-shot. Retrieval quality matters at least as much as the number of shots: 1-shot highest chrF++ improves All Tags by 45.4 points, essentially matching 5-shot Random. More generally, high-similarity retrieval consistently outperforms low-similarity selection, and this gap persists as $k$ increases. The best dev configuration is highest chrF++ with $k{=}10$, which improves All Tags by 63.9 points over zero-shot. We therefore use $(k^\ast,m^\ast)=(10,\texttt{highest chrF++})$ as the \emph{best-ICL} setting for test evaluation.
.

\paragraph{Test Results}
Table~\ref{tab:mtag_test} shows that the selected ICL configuration produces large gains on All Tags for all LLMs. The largest absolute improvements are observed for the open-weight models, with gains of 62.6 points for Qwen3 and 49.7 points for Llama4, while the proprietary models improve by 46.5 and 37.8 points for GPT5.2 and Gemini3, respectively. These gains substantially narrow the gap to the supervised baseline in Tag F\textsubscript{1}, but All Tags remains lower than CAMeL Tools, confirming that exact full-bundle prediction is 
considerably harder for LLMs than feature-level accuracy alone.




\begin{table}[t]

\centering
\small
\setlength{\tabcolsep}{5pt}
\begin{tabular}{lcccc}
\toprule
 & \multicolumn{2}{c}{\textbf{Zero-shot}} & \multicolumn{2}{c}{\textbf{Best-ICL}} \\
\cmidrule(lr){2-3}\cmidrule(lr){4-5}
\textbf{Model} &\textbf{ All Tags} & \textbf{Tag F\textsubscript{1}} & \textbf{All Tags} & \textbf{Tag F\textsubscript{1}} \\
\midrule
Llama4   & 7.7  & 78.9 & 57.4 & 90.1 \\
Qwen3    & 13.9 & 83.1 & 76.5 & 97.0 \\
GPT5.2   & 39.7 & 89.9 & 86.2 & 98.4 \\
Gemini3  & 48.8 & 93.0 & 86.6 & 98.5 \\
\midrule
CAMeL Tools & \textbf{96.3} & \textbf{99.6} & -- & -- \\
\bottomrule
\end{tabular}
\caption{\textbf{Morphosyntactic Tagging} results on PATB Test. Best-ICL uses the Dev-selected configuration ($k{=}10$; highest chrF++ retrieval).}
\label{tab:mtag_test}
\end{table}
\begin{table}[t]
\centering
\small
\tabcolsep12pt
\begin{tabular}{lccc}
\toprule
\textbf{Model} & \textbf{LS} & \textbf{UAS} & \textbf{LAS} \\
\midrule
Llama4   & 62.1 & 41.8 & 28.4 \\
Qwen3    & 62.2 & 42.5 & 28.9 \\
GPT5.2   & 82.1 & 84.1 & 73.1 \\
Gemini3  & 87.0 & 85.2 & 78.8 \\
\midrule
CamelParser2.0 & 93.6 & 89.7 & 87.5 \\
CamelParser2.1 & \textbf{94.4} & \textbf{90.6} & \textbf{88.4} \\
\bottomrule
\end{tabular}
\caption{\textbf{Dependency Parsing} results on CAMeLTB Dev under gold tokenization (zero-shot).}
\label{tab:parse_dev_zeroshot}

\end{table}



\begin{table*}[t!]
\centering
\tabcolsep3pt
\small
\begin{tabular}{lccccccccccccccc}
\toprule
\textbf{Number of Shots} 
& \multicolumn{3}{c}{\textbf{0}} 
& \multicolumn{3}{c}{\textbf{1}} 
& \multicolumn{3}{c}{\textbf{3}} 
& \multicolumn{3}{c}{\textbf{5}} 
& \multicolumn{3}{c}{\textbf{10}} \\
\cmidrule(lr){2-4} \cmidrule(lr){5-7} \cmidrule(lr){8-10} \cmidrule(lr){11-13} \cmidrule(lr){14-16}
\textbf{Selection Method} 
& \textbf{LS} & \textbf{UAS} & \textbf{LAS}
& \textbf{LS} & \textbf{UAS} & \textbf{LAS}
& \textbf{LS} & \textbf{UAS} & \textbf{LAS}
& \textbf{LS} & \textbf{UAS} & \textbf{LAS}
& \textbf{LS} & \textbf{UAS} & \textbf{LAS} \\
\midrule
Lowest Cosine Similarity  & --   & --   & --   & 63.1 & 43.7 & 30.5 & 64.9 & 45.6 & 33.3 & 65.7 & 46.2 & 34.6 & 66.6 & 47.7 & 36.4 \\
Lowest chrF++             & --   & --   & --   & 62.6 & 43.2 & 29.7 & 64.9 & 43.9 & 32.5 & 65.4 & 44.6 & 33.4 & 65.5 & 45.4 & 34.2 \\
Random                    & 62.2 & 42.5 & 28.9 & 71.8 & 55.7 & 46.3 & 76.1 & 62.6 & 54.7 & 77.4 & 65.5 & 58.0 & 80.0 & 69.0 & 62.2 \\
Highest Cosine Similarity & --   & --   & --   & 78.6 & 66.7 & 59.2 & 82.6 & 73.0 & 66.7 & 84.5 & 75.3 & 69.9 & 85.8 & 77.6 & 72.5 \\
Highest chrF++            & --   & --   & --   & \textbf{79.8} & \textbf{68.8} & \textbf{61.5} & \textbf{84.2} & \textbf{74.5} & \textbf{69.0} & \textbf{85.4} & \textbf{76.5} & \textbf{71.4} & \textbf{86.9} & \textbf{78.5} & \textbf{74.0} \\
\bottomrule
\end{tabular}
\caption{\textbf{Dependency Parsing} performance comparison of shot selection methods across different numbers of shots for \textbf{Qwen3} on CAMeLTB Dev under gold tokenization. The $k{=}0$ (zero-shot) results correspond to the \textit{Random} row, since no retrieval is performed in that setting.}
\label{tab:parse_selection_methods}
\end{table*}




\begin{table*}[t!]
\centering
\small
\setlength{\tabcolsep}{5pt}
\begin{tabular}{lcccccccc}
\toprule
 & \multicolumn{4}{c}{\textbf{Zero-shot}} & \multicolumn{4}{c}{\textbf{Best-ICL}} \\
\cmidrule(lr){2-5}\cmidrule(lr){6-9}
\textbf{Model} & \textbf{Tok F\textsubscript{1}} & \textbf{LS} & \textbf{UAS} & \textbf{LAS} & \textbf{Tok F\textsubscript{1}} & \textbf{LS} & \textbf{UAS} & \textbf{LAS} \\
\midrule
Llama4    &100.0 & 60.8 & 43.0 & 28.3 &100.0 & 82.9 & 73.7 & 67.4 \\
Qwen3     &100.0 & 60.6 & 40.8 & 26.9 &100.0 & 85.7 & 79.3 & 73.8 \\
GPT5.2    &100.0 & 81.8 & 84.1 & 72.8 &100.0 & 91.8 & 88.6 & 85.0 \\
Gemini3   &100.0 & 86.5 & 85.1 & 78.3 &100.0 & 94.0 & 90.7 & 88.2 \\
\midrule
CamelParser2.0 &100.0 & 93.1 & 89.8 & 87.5 &-- & -- & -- & -- \\
CamelParser2.1 &100.0 & \textbf{94.6} & \textbf{91.7} & \textbf{89.3} &-- & -- & -- & -- \\
\bottomrule
\end{tabular}
\caption{\textbf{Dependency Parsing} results on CAMeLTB Test (gold tokenization). Best-ICL uses the Dev-selected configuration ($k{=}10$; highest chrF++ retrieval).}
\label{tab:parse_test_gold}
\vspace{10pt}

\centering
\small
\setlength{\tabcolsep}{5pt}
\begin{tabular}{lcccccccc}
\toprule
 & \multicolumn{4}{c}{\textbf{Zero-shot}} & \multicolumn{4}{c}{\textbf{Best-ICL}} \\
\cmidrule(lr){2-5}\cmidrule(lr){6-9}
\textbf{Model} & \textbf{Tok F\textsubscript{1}} & \textbf{LS} & \textbf{UAS} & \textbf{LAS} & \textbf{Tok F\textsubscript{1}} & \textbf{LS} & \textbf{UAS} & \textbf{LAS} \\
\midrule
Llama4   & 49.5 & 42.7 & 37.6 & 23.8 & 87.8 & 78.6 & 71.9 & 66.0 \\
Qwen3    & 51.2 & 49.1 & 37.4 & 24.1 & 84.9 & 80.6 & 74.1 & 68.8 \\
GPT5.2   & 93.7 & 70.5 & 69.7 & 60.1 & \textbf{98.2} & 90.9 & 87.4 & 84.1 \\
Gemini3  & 92.8 & 78.1 & 75.0 & 68.4 & 96.2 & 91.9 & 88.1 & 85.3 \\
\midrule
CamelParser2.0 & 98.0 & 91.5 & 87.3 & 84.8 & -- & -- & -- & -- \\
CamelParser2.1 & 98.0 & \textbf{92.5} & \textbf{88.5} & \textbf{86.0} & -- & -- & -- & -- \\
\bottomrule
\end{tabular}
\caption{\textbf{Dependency Parsing} results on CAMeLTB Test (raw text). Best-ICL uses the Dev-selected configuration ($k{=}10$; highest chrF++ retrieval).}
\label{tab:parse_test_raw}
\end{table*}


\subsection{Dependency Parsing}
\label{subsec:results_parse}

\paragraph{Baseline}
Table~\ref{tab:parse_dev_zeroshot} reports zero-shot dev results under gold tokenization. The supervised baselines remain strongest, with CamelParser2.1 improving over CamelParser2.0 through multiroot decoding. Gemini3 is the best zero-shot LLM and Qwen3 is the strongest open-weight model. The large gap between proprietary and open-weight models in zero-shot highlights the difficulty of structured Arabic parsing without task-specific supervision. We therefore select Qwen3 for ICL analysis.

\paragraph{ICL}
Table~\ref{tab:parse_selection_methods} shows that ICL substantially improves parsing performance, but gains depend strongly on demonstration selection. With random 1-shot prompting, LAS improves by 17.4 points over zero-shot, indicating that even a single example helps adapt the model to the task. However, retrieval quality is crucial: at $k{=}1$, highest chrF++ improves LAS by 32.6 points, far exceeding low-similarity retrieval, which yields only marginal gains. This advantage persists as $k$ increases, and highest chrF++ remains the strongest method throughout, reaching a total LAS improvement of 45.1 points at $k{=}10$. We therefore select $(k^\ast,m^\ast)=(10,\texttt{highest chrF++})$ as the \emph{best-ICL} setting for test evaluation.

\paragraph{Gold Tokenization Test Results}
Under gold tokenization (Table~\ref{tab:parse_test_gold}), best-ICL yields large gains for all LLMs, with LAS improvements ranging from 9.9 to 46.9 points. The largest gains are observed for the open-weight models, especially Qwen3 (+46.9) and Llama4 (+39.1), while the proprietary models improve more modestly from stronger zero-shot baselines. Gemini3 under best-ICL slightly surpasses CamelParser2.0 (88.2 vs.\ 87.5 LAS), but CamelParser2.1 is stronger at 89.3 LAS.

\paragraph{Raw Text Test Results}
In the raw-text setting (Table~\ref{tab:parse_test_raw}), tokenization is a major source of zero-shot degradation, especially for open-weight models. Best-ICL substantially improves both tokenization and parsing, yielding Tok-F\textsubscript{1} gains of 33.7 and 38.3 points and LAS gains of 44.7 and 42.2 points for Qwen3 and Llama4, respectively. The proprietary models are considerably more robust in zero-shot, but still benefit from ICL, with smaller tokenization gains and LAS improvements of 24.0 points for GPT5.2 and 16.9 for Gemini3. Multiroot decoding raises the supervised raw-text LAS from 84.8 for CamelParser2.0 to 86.0 for CamelParser2.1. Under best-ICL, Gemini3 surpasses CamelParser2.0 on LS, UAS, and LAS, but remains below CamelParser2.1; GPT5.2 also remains competitive with the supervised parsers. Overall, these results show that ICL improves not only dependency prediction but also 
tokenization decisions. 

\subsection{Joint PATB--CATiB Evaluation}
\label{subsec:joint_patb}

Table~\ref{tab:joint_patb} tests whether supplying a model's predicted morphology improves its raw-text parsing relative to performing the two tasks independently.

\begin{table*}[t!]
\centering
\small
\setlength{\tabcolsep}{4pt}
\begin{tabular}{llcccccccccc}
\toprule
\multirow{2}{*}{\textbf{System}} & \multirow{2}{*}{\textbf{Setting}}
 & \multirow{2}{*}{\textbf{All Tags}} & \multirow{2}{*}{\textbf{Tag F\textsubscript{1}}}
 & \multicolumn{4}{c}{\textbf{Independent parsing}}
 & \multicolumn{4}{c}{\textbf{Cascaded parsing}} \\
\cmidrule(lr){5-8}\cmidrule(lr){9-12}
 & & &
 & \textbf{Tok F\textsubscript{1}} & \textbf{LS} & \textbf{UAS} & \textbf{LAS}
 & \textbf{Tok F\textsubscript{1}} & \textbf{LS} & \textbf{UAS} & \textbf{LAS} \\
\midrule
Llama4  & Zero-shot & 7.7  & 78.9 & 70.9 & 44.4 & 41.1 & \underline{24.9} & 73.4 & 43.8 & 40.2 & 24.4 \\
Llama4  & Best-ICL  & 57.4 & 90.1 & 86.2 & 74.2 & 69.1 & \underline{60.8} & 85.7 & 73.4 & 66.4 & 58.6 \\
\addlinespace[2pt]
Qwen3   & Zero-shot & 13.9 & 83.1 & 73.2 & 50.6 & 42.4 & \underline{27.7} & 74.2 & 46.2 & 35.6 & 23.7 \\
Qwen3   & Best-ICL  & 76.5 & 97.0 & 86.2 & 79.2 & 73.5 & 67.8 & 88.1 & 80.0 & 73.8 & \underline{68.1} \\
\addlinespace[2pt]
GPT5.2  & Zero-shot & 39.7 & 89.9 & 90.1 & 75.0 & 59.9 & 52.3 & 93.1 & 77.6 & 64.4 & \underline{57.4} \\
GPT5.2  & Best-ICL  & 86.2 & 98.4 & 98.0 & 90.4 & 82.5 & 78.4 & 98.4 & 92.0 & 83.7 & \underline{80.7} \\
\addlinespace[2pt]
Gemini3 & Zero-shot & 48.8 & 93.0 & 95.7 & 83.6 & 79.1 & 72.1 & 94.0 & 84.2 & 80.0 & \underline{72.7} \\
Gemini3 & Best-ICL  & 86.6 & 98.5 & 96.0 & 93.7 & 88.1 & 85.2 & 95.4 & 94.1 & 88.4 & \underline{85.8} \\
\midrule
CamelParser2.0 & Supervised & \textbf{96.3} & \textbf{99.6} & \textbf{98.7} & 96.7 & 92.1 & \underline{90.9} & \textbf{98.7} & 96.7 & 92.1 & \underline{90.9} \\
CamelParser2.1 & Supervised & \textbf{96.3} & \textbf{99.6} & \textbf{98.7} & \textbf{97.2} & \textbf{92.6} & \underline{\textbf{91.5}} & \textbf{98.7} & \textbf{97.2} & \textbf{92.6} & \underline{\textbf{91.5}} \\
\bottomrule
\end{tabular}
\caption{\textbf{Joint PATB--CATiB results.} LLM parsing is performed independently or with the same model's predicted morphology. Morphology scores are shared across conditions; CamelParser scores are repeated because CamelParser already consumes the morphological analyses produced by CAMeL Tools. Bold indicates the best score in each column. Within each row, the higher LAS between independent and cascaded parsing is underlined.}
\label{tab:joint_patb}
\end{table*}

The row-wise comparison shows that the two strongest LLMs, GPT5.2 and Gemini3, consistently achieve higher LAS when their predicted morphology is supplied to the parser, in both zero-shot and Best-ICL settings. Qwen3 also favors cascading under Best-ICL, whereas Llama4 favors independent parsing. Thus, predicted morphology is most useful for the stronger parsers, although its benefit is not universal.

Retrieval-based ICL provides the more consistent gain: every LLM improves from zero-shot to Best-ICL under both independent and cascaded parsing. Gemini3 is the strongest cascaded LLM under Best-ICL, while CamelParser2.1 remains the strongest system overall. Together, these results show that retrieved demonstrations substantially strengthen parsing and that cascading morphology gives the best-performing LLMs an additional LAS improvement.

\section{Analysis and Discussion}
\label{sec:analysis}
\subsection{Cost Comparison: Time, Money, and Data}



While Gemini3 achieves competitive performance, it incurs substantially higher costs than Camel tools and CamelParser. It is 1{,}000$\times$–8{,}000$\times$ slower\footnote{All open-weight and supervised experiments were run on a single node with 4 A100 GPUs.} and costs over \$200 in prompting on the test sets alone, whereas Camel tools are free and significantly faster (Appendix~\ref{app:time-money}).
Moreover, retrieval-based ICL depends on extensive annotated data: the best tagging setup covers 51\% of PATB sentences (61\% of tokens), and parsing requires 67\% (71\% of tokens) (Appendix~\ref{app:data-req}). 
Overall, LLM-based gains come at significant cost in time, money, and reliance on pre-annotated data.

\subsection{Morphosyntactic Tagging}






On the PATB Test split, we compare per-feature accuracy between CAMeL~Tools and Gemini3 under Best-ICL (Appendix~\ref{app:mtag_feature_accuracy}). Most features show comparable performance, with notable differences in form gender, form number, and case: \texttt{gen} (3.4\%), \texttt{num} (3.9\%), and \texttt{cas} (3.1\%).

To better understand these gaps, we analyze representative examples. For form gender and form number, most discrepancies arise from mismatches between surface-form and functional features in Arabic \citep{alkuhlani-habash-2011-corpus}. For instance, broken plurals such as \<طلبة> \textit{Talaba{\TAMAR}} `students' are feminine singular in form but functionally masculine plural. Gemini3 often predicts the correct functional interpretation, but is penalized under the form-feature annotation scheme. While this reflects a limitation of the evaluation setup, it is consistent with prior work \cite{zalmout-etal-2018-noise,inoue-etal-2022-morphosyntactic}.
For case (\texttt{cas}), the main difficulty lies in the \texttt{u} (undefined) category, which appears in the reference for many proper nouns and indeclinable nouns and adjectives lacking overt case marking, e.g. \<أخرى> \textit{{\AHAMZAUP}uxra{\AMAQ}} `other'. Gemini3 achieves 72.6\% accuracy on \texttt{cas=u}, compared to 94.4\% for CAMeL~Tools. Excluding \texttt{cas=u}, performance becomes nearly identical (98.6\% vs.\ 98.4\%), indicating that most of the overall difference is concentrated in this category. 

These observations suggest that  differences between CAMeL~Tools and Gemini3 are smaller than they initially appear and largely reflect specifics of the reference annotation and task definition.

\subsection{Dependency Parsing}
\paragraph{Error Patterns}
 


On the CAMeLTB Test set with gold tokenization, root attachment explains the ordering between Gemini3 and CamelParser2.0. The gold trees average 1.41 roots per input, with 545 of 1{,}918 inputs containing multiple roots, which CamelParser2.0's exactly-one-root decoder cannot reproduce. Gemini3 consequently has much higher head accuracy on tokens attached to the gold root (85.5\% vs.\ 66.1\%). CamelParser2.1 keeps the same checkpoint and scoring model but uses projective multiroot decoding, raising gold-root head accuracy to 89.5\% and overall LAS to 89.3\% while leaving non-root LAS unchanged at 89.3\%.

\paragraph{Differences by Genre and Length}


The genre breakdown in Appendix~\ref{app:genre-analysis} shows that much of the variation follows root structure. CamelParser2.0 outperforms Gemini3 on 6 of 13 CAMeLTB genres, whereas CamelParser2.1 does so on 10; multiroot decoding improves 11 genres, with the largest gains on Quran, OT, NT, and Sara. Importantly, Gemini3 is strongest on Classical Arabic overall, leading on Odes, Quran, and Hayy, whereas CamelParser2.1 is strongest on MSA overall. By length, Gemini3 leads on short inputs, while CamelParser2.1 leads on mid-length and long inputs. Thus, multiroot decoding explains much of the cross-genre variation, but the results also reveal a broader contrast between Gemini3's strength on Classical Arabic and the supervised parser's strength on MSA.

\subsection{Tokenization Error Analysis}

To analyze tokenization on the CAMeLTB Test split, we manually examined a 500-sentence sample (6,297 gold tokens) from the raw-text setting, aligning CAMeL~Tools and Gemini3 Best-ICL outputs with the gold token sequence. Unlike earlier evaluations that ignored normalizable differences, this analysis accounts for all discrepancies, explaining Gemini3’s lower tokenization scores.
As shown in Appendix~\ref{app:tokerr} Table~\ref{tab:tok-errors}, the systems exhibit distinct error profiles. CamelParser uses CAMeL Tools for tokenization, which is trained on PATB (MSA newswire) and have limited support for Classical Arabic and other CAMeLTB genres. The CAMeL~Tools errors are dominated by punctuation normalization (55\%), e.g., Western vs.\ Arabic punctuation (``?'' vs.\ \<؟>), which rarely affect token boundaries, along with moderate under-tokenization (11\%) and normalization differences (15\%). In contrast, Gemini3’s errors are primarily hallucinations (56\%), generating tokens not present in the input (e.g., English words or unrelated tokens such as \textit{plus}, \textit{index}, \textit{standard}, \textit{village}, and \textit{Current}, as well as occasional non-Arabic characters such as  
\includegraphics[]{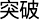} \textit{tūpò} `breakthrough').
Excluding hallucinations, Gemini3 shows relatively few tokenization errors.
Both systems also exhibit minor discrepancies due to gold annotation issues and normalization variation. Overall, CamelParser errors are largely superficial, while Gemini3’s lower performance is mainly driven by hallucinated tokens rather than systematic tokenization errors.  
This analysis highlights some of the weaknesses of the baseline models as well as the LLM models, all of which open interesting pathways for future research.

\section{Conclusion and Future Work}
\label{sec:conclusion}
\textcolor{black}{
We evaluated instruction-tuned LLMs on Arabic morphosyntactic tagging and dependency parsing, both independently and in a cascade, under zero-shot and retrieval-based ICL. Relevant demonstrations substantially improve both tasks, especially for open-weight models. Strong LLMs approach supervised systems on individual morphological features, although exact feature bundles remain difficult, and become competitive with specialized dependency parsers. The strongest LLM configuration reaches 85.8 LAS, but raw-text tokenization remains an important source of error. Overall, LLMs can produce useful explicit analyses of Arabic morphology and syntax, although competitive performance relies on substantial annotated data and computation. Future work should explore joint modeling of tokenization, morphology, and syntax and evaluate these methods on other Arabic dialects.
}

\section*{Limitations}
Our study has three main limitations. First, the evaluation focuses on Standard Arabic and does not cover dialectal varieties, so the findings may not generalize across Arabic. Second, retrieval-based ICL is computationally and financially costly, and its long prompts may be impractical for models with smaller context windows. Finally, although our cascade connects morphology and parsing, it does not jointly model tokenization, morphology, and syntax. The independent PATB--CATiB comparison is therefore a pipeline-level ablation: morphology results are shared across conditions, which differ only in whether those predictions are supplied to the parser.

\section*{Ethics Statement}
This work evaluates large language models (LLMs) on structured linguistic analysis tasks for Standard Arabic. We use publicly available Arabic treebanks and do not introduce new annotated data involving human subjects. As such, risks related to privacy, consent, or personal data are minimal.

However, LLMs may reflect biases present in their training data, including dialectal, regional, or stylistic biases in Arabic. Our evaluation focuses on  Standard Arabic and may not generalize to dialectal varieties, potentially reinforcing disparities in language technology coverage. In addition, reliance on proprietary models raises concerns about reproducibility and transparency.

We emphasize that our findings are diagnostic and should not be interpreted as evidence of robust linguistic competence. Misinterpretation of model outputs in downstream applications—especially in educational or linguistic contexts—may lead to incorrect analyses. We encourage careful use of LLM-generated linguistic annotations and further research on fairness and robustness across Arabic varieties.

We used AI writing assistance within the scope of ``Assistance
purely with the language of the paper'' described in the ACL Policy on Publication Ethics.

\section*{Acknowledgments}
We thank Go Inoue and Muhammed AbuOdeh for insightful discussions and feedback. We gratefully acknowledge the support and computational resources provided by the High Performance Computing Center at New York University Abu Dhabi, which supported the model training and experiments in this work. This research is in part with support from Google.org and the Google Cloud Research Credits program for the Gemini Academic Program.


\bibliography{custom,camel-bib-v3,anthology-1,anthology-2}

\newpage
\appendix
\raggedbottom
\section{Prompts}
\subsection{Morphosyntactic Tagging Prompt}
\label{app:mtag_prompt}
The panel labels below are editorial and decompose the fixed system prompt into
functional parts for readability. 
The prompt content and ordering are preserved.

\begin{promptpart}{Task, tokens, and output} 
\begin{alltt}\footnotesize
You are an expert Arabic morphosyntactic
analyzer trained on the Penn Arabic
Treebank (PATB).

Your task is to analyze the Arabic
sentences and produce a complete
morphosyntactic analysis for EACH TOKEN.
\end{alltt}
\begin{alltt}\footnotesize
TOKENIZATION

- The sentence is already tokenized and
  tokens are separated by spaces.
- Do NOT merge or split tokens.
- Analyze tokens strictly in the given
  order.
\end{alltt}
\begin{alltt}\footnotesize
OUTPUT REQUIREMENTS

- Output MUST be valid JSON.
- Output MUST contain a single key:
  "tokens".
- "tokens" is a list of token objects.
- Each token object MUST include:
    - "word": the surface form exactly
      as given
    - "analysis": a full feature
      structure
- ALL features listed below MUST appear
  for every token.
- Use ONLY the allowed values.
- Do NOT include explanations, comments,
  or extra fields in the output.
\end{alltt}
\end{promptpart}

\begin{promptpart}{Morphosyntactic feature inventory}
\noindent{\footnotesize\ttfamily
MORPHOLOGICAL \& SYNTACTIC FEATURES\par
(WITH ALLOWED VALUES \& DESCRIPTIONS)}
\par\smallskip
\begin{alltt}\footnotesize
asp - (Aspect)
- c : Command
- i : Imperfective
- p : Perfective
- na : Not applicable

cas - (Case)
- n : Nominative
- a : Accusative
- g : Genitive
- na : Not applicable
- u : Undefined

form_gen - (Form gender)
- f : Feminine
- m : Masculine
- na : Not applicable

form_num - (Form number)
- s : Singular
- d : Dual
- p : Plural
- na : Not applicable
- u : Undefined

\end{alltt}
\begin{alltt}\footnotesize
mod - (Mood)
- i : Indicative
- j : Jussive
- s : Subjunctive
- na : Not applicable
- u : Undefined

per - (Person)
- 1 : 1st
- 2 : 2nd
- 3 : 3rd
- na : Not applicable

stt - (State)
- c : Construct/Poss/Idafa
- d : Definite
- i : Indefinite
- na : Not applicable
- u : Undefined

vox - (Voice)
- a : Active
- p : Passive
- na : Not applicable
- u : Undefined
\end{alltt}
\end{promptpart}

\begin{promptpart}{Part-of-speech feature inventory}
\begin{alltt}\footnotesize
pos - (Part of speech)
- noun : Noun
- noun_prop : Proper noun
- noun_num : Number noun
- noun_quant : Quantity noun
- adj : Adjective
- adj_comp : Comparitive adjective
- adj_num : Numerical adjective
- adv : Adverb
- adv_interrog : Interrogative adverb
- adv_rel : Relative adverb
\end{alltt}
\begin{alltt}\footnotesize
- pron : Pronoun
- pron_dem : Demonstrative pronoun
- pron_exclam : Pronoun exclamation
- pron_interrog : Interrogative pronoun
- pron_rel : Relative pronoun
- verb : Verb
- verb_pseudo : Pseudo verb
- part : Particle
- part_dem : Demonstrative particle
- part_det : Determiner particle
- part_focus : Focus particle
- part_fut : Future marker particle
- part_interrog : Interrogative particle
- part_neg : Negative particle
- part_restrict : Restrictive particle
- part_verb : Verbal particle
- part_voc : Vocalized particle
- prep : Preposition
- abbrev : Abbreviation
- punc : Punctuation
- conj : Conjunction
- conj_sub : Subordinating conjunction
- interj : Interjection
- digit : Digital numbers
- latin : Latin/foreign
\end{alltt}
\end{promptpart}

\begin{promptpart}{Clitic feature inventory}
\noindent\textbf{Proclitics}\par
\begin{alltt}\footnotesize
prc0 - (Article proclitic)
- 0 : No proclitic
- na : Not applicable
- Aa_prondem : Demonstrative particle Aa
- Al_det : Determiner
- AlmA_neg : Determiner Al + negative
  particle mA
- lA_neg : Negative particle lA
- mA_neg : Negative particle mA
- ma_neg : Negative particle ma
- mA_part : Particle mA
- mA_rel : Relative pronoun mA

prc1 - (Preposition proclitic)
- 0 : No proclitic
- na : Not applicable
- <i\$_interrog : Interrogative ish
- bi_part : Particle bi
- bi_prep : Preposition bi
- bi_prog : Progressive verb particle bi
- Ea_prep : Preposition Ea
- EalaY_prep : Preposition EalaY
- fiy_prep : Preposition fy
- hA_dem : Demonstrative hA
- Ha_fut : Future marker Ha
- ka_prep : Preposition ka
- la_emph : Emphatic particle la
- la_prep : Preposition la
- la_rc : Response conditional la
- libi_prep : Preposition li +
  preposition bi
- laHa_emphfut : Emphatic la + future
  marker Ha
- laHa_rcfut : Response conditional la +
  future marker Ha
\end{alltt}
\begin{alltt}\footnotesize
- li_jus : Jussive li
- li_sub : Subjunctive li
- li_prep : Preposition li
- min_prep : Preposition min
- sa_fut : Future marker sa
- ta_prep : Preposition ta
- wa_part : Particle wa
- wa_prep : Preposition wa
- wA_voc : Vocative wA
- yA_voc : Vocative yA

prc2 - (Conjunction proclitic)
- 0 : No proclitic
- na : Not applicable
- fa_conj : Conjunction fa
- fa_conn : Connective particle fa
- fa_rc : Responsive conditional fa
- fa_sub : Subordinating conjunction fa
- wa_conj : conjunction wa
- wa_part : particle wa
- wa_sub : Subordinating conjunction wa

prc3 - (Question proclitic)
- 0 : No proclitic
- na : Not applicable
- >a_ques : Interrogative partical >a
\end{alltt}
\end{promptpart}

\begin{promptpart}{Clitic feature inventory: enclitics}
\begin{alltt}\footnotesize
enc0 - (Pronominal enclitic)
- 0 : No enclitic
- na : Not applicable
- 1s_dobj : 1st person singular direct
  object
- 1s_poss : 1st person singular
  possessive
- 1s_pron : 1st person singular pronoun
- 1p_dobj : 1st person plural direct
  object
- 1p_poss : 1st person plural possessive
- 1p_pron : 1st person plural pronoun
- 2d_dobj : 2nd person dual direct object
- 2d_poss : 2nd person dual possessive
- 2d_pron : 2nd person dual pronoun
- 2p_dobj : 2nd person plural direct
  object
- 2p_poss : 2nd person plural possessive
- 2p_pron : 2nd person plural pronoun
- 2fs_dobj : 2nd person feminine singular
  direct object
- 2fs_poss : 2nd person feminine singular
  possessive
- 2fs_pron : 2nd person feminine singular
  pronoun
- 2fp_dobj : 2nd person feminine plural
  direct object
- 2fp_poss : 2nd person feminine plural
  possessive
- 2fp_pron : 2nd person feminine plural
  pronoun
- 2ms_dobj : 2nd person masculine singular
  direct object
- 2ms_poss : 2nd person masculine singular
  possessive
- 2ms_pron : 2nd person masculine singular
  pronoun
- 2mp_dobj : 2nd person masculine plural
  direct object
- 2mp_poss : 2nd person masculine plural
  possessive
- 2mp_pron : 2nd person masculine plural
  pronoun
\end{alltt}
\end{promptpart}

\begin{promptpart}{Clitic feature inventory: enclitics (continued)}
\begin{alltt}\footnotesize
- 3d_dobj : 3rd person dual direct object
- 3d_poss : 3rd person dual possessive
- 3d_pron : 3rd person dual pronoun
- 3p_dobj : 3rd person plural direct
  object
- 3p_poss : 3rd person plural possessive
- 3p_pron : 3rd person plural pronoun
- 3fs_dobj : 3rd person feminine singular
  direct object
- 3fs_poss : 3rd person feminine singular
  possessive
- 3fs_pron : 3rd person feminine singular
  pronoun
- 3fp_dobj : 3rd person feminine plural
  direct object
- 3fp_poss : 3rd person feminine plural
  possessive
- 3fp_pron : 3rd person feminine plural
  pronoun
- 3ms_dobj : 3rd person masculine singular
  direct object
- 3ms_poss : 3rd person masculine singular
  possessive
- 3ms_pron : 3rd person masculine singular
  pronoun
- 3mp_dobj : 3rd person masculine plural
  direct object
- 3mp_poss : 3rd person masculine plural
  possessive
- 3mp_pron : 3rd person masculine plural
  pronoun
- Ah_voc : Vocative particle Ah
- lA_neg : Negative particle lA
- ma_interrog : Interrogative pronoun ma
- mA_interrog : Interrogative pronoun mA
- man_interrog : Interrogative pronoun
  man
- ma_rel : Relative pronoun ma
- mA_rel : Relative pronoun mA
- man_rel : Relative pronoun man
- ma_sub : Subordinating conjunction ma
- mA_sub : Subordinating conjunction mA
\end{alltt}
\end{promptpart}

\begin{promptpart}{Query}
\begin{alltt}\footnotesize
Now analyze the following Arabic sentence
with full PATB-style morphosyntax.
Tokens:
["TOKEN_1", "TOKEN_2", ..., "TOKEN_n"]
\end{alltt}
\end{promptpart}

\newpage
\subsection{Dependency Parsing Prompts}
\label{app:parsing_prompts}
The panel labels below are editorial and expose the functional structure of
the prompts. The prompt content and ordering are preserved.

\subsubsection{Gold-tokenization prompt}
The following prompt was used when gold CATiB tokenization was provided to
the model.

\begin{promptpart}{Task, input, and output}
\begin{alltt}\footnotesize
You are an expert Arabic dependency
annotator trained on the Columbia Arabic
Treebank (CATiB)
conventions.
Your task is to annotate Arabic
sentences using the official CATiB
methodology.
\end{alltt}
\begin{alltt}\footnotesize
Input format
You are given a single Arabic sentence
already tokenized according to CATiB
conventions.
The provided token sequence must be used
exactly as-is for dependency annotation.
\end{alltt}
\begin{alltt}\footnotesize
Output format
Output only a JSON object with a single
key "parses".
The value of "parses" must be an array of
objects, one per token, each containing:
- id: integer (1-based token index)
- form: identical to the produced token
  (including "+" markers)
- head: integer (1-based index of the
  syntactic head; 0 = root)
- deprel: dependency relation label
Each non-root token must have exactly
one head and one dependency relation.
No explanations, commentary, or
additional formatting should be included.
\end{alltt}
\end{promptpart}

\begin{promptpart}{Relations and attachment rules}
\begin{alltt}\footnotesize
Dependency relations
SBJ - Subject (verbal, nominal, or
      copular sentence)
OBJ - Object of verbs, deverbal nouns,
      prepositions, and conjunctions
PRD - Predicate of incomplete verbs or
      verb-like particles
TPC - Topic in a complex nominal sentence
      with a different internal subject
IDF - Idafa (genitive possessor)
TMZ - Tamyiz / specification
MOD - Modifier (adjectival, adverbial,
      prepositional, appositional,
      relative,
      demonstrative, etc.)
--- - Flat / multiword proper-name
      connector
\end{alltt}
\begin{alltt}\footnotesize
Sentence and clause structure
Verbal sentences
- Verb is the clause head.
- SBJ marks explicit subjects (pre- or
  post-verbal).
- OBJ marks direct or pronominal objects.
- Passive verbs take surface SBJ for the
  underlying object.
- Verbal particles (\AR{قد}, \AR{سوف},
  \AR{لم}, \AR{لن}) attach as MOD.
Nominal sentences
- Predicate heads the sentence.
- Topic is SBJ.
- If preceded by an incomplete or
  verb-like particle,
  the particle heads both SBJ and PRD.
Complex sentences
- If topic and embedded subject are the
  same, use SBJ.
- Otherwise, use TPC.
\end{alltt}
\end{promptpart}

\begin{promptpart}{Phrase-level and attachment rules}
\begin{alltt}\footnotesize
Prepositional phrases
- Preposition heads its OBJ.
- Preposition attaches to its governor
  as MOD.
Nominal modification
- IDF marks possessors in idafa (only
  one IDF per head).
- MOD marks adjectives, relative clauses,
  demonstratives, and apposition.
- TMZ marks specification.
Relative clauses
- Attach to head noun using MOD.
- Relative pronoun heads the clause when
  present.
\end{alltt}
\begin{alltt}\footnotesize
Coordination
- First conjunct attaches to conjunction
  as MOD.
- Conjunction attaches to second conjunct
  as OBJ.
- Sentence-initial \AR{و} attaches as MOD
  to the following clause head.
Subordination
- Main clause head attaches to
  subordinator as MOD.
- Subordinate clause head attaches to
  subordinator as OBJ.
- \AR{أن} / \AR{أنّ} may attach by
  grammatical role (SBJ, PRD, etc.).
Punctuation
- Punctuation attaches as MOD to the
  governing node.
- Sentence-final punctuation attaches to
  the main clause head.
\end{alltt}
\end{promptpart}

\begin{promptpart}{Query}
\begin{alltt}\footnotesize
Now annotate the following Arabic
sentence using CATiB.
Tokens:
["TOKEN_1", "TOKEN_2", ..., "TOKEN_n"]
\end{alltt}
\end{promptpart}

\newpage
\subsubsection{Raw-text additions}
For the raw-text setting, the same prompt shown above was used, with the
following blocks prepended to the beginning of the prompt.

\begin{promptpart}{Raw-text tokenization additions}
\textbf{Input and processing order}
\begin{alltt}\footnotesize
Input format
You are given a single Arabic sentence as
RAW TEXT (a normal string), not
pre-tokenized.
Tokenization (YOU MUST DO THIS FIRST)
Words in CATiB are whitespace- and
punctuation-separated strings.
For annotation purposes, words must be
broken into tokens following CATiB
conventions.
All diacritics must be removed.
\end{alltt}
\textbf{Proclitic inventory}
\begin{alltt}\footnotesize
Only the following clitics are separated
from the word.
Proclitics
(written as separate tokens ending with
"+"):

- \AR{أ}+ question particle
- \AR{و}+ and
- \AR{ف}+ so / then
- \AR{ل}+ for / to / so that /
  emphatic particle
- \AR{ب}+ by / with
- \AR{ك}+ as / like
\end{alltt}
\textbf{Enclitic inventory}
\begin{alltt}\footnotesize
Enclitics
(written as separate tokens beginning
with "+"):

- pronominal object or possessive clitics
  such as
+\AR{ه} him / his
+\AR{ها} her
+\AR{هم} them / their
+\AR{كما} you (dual)
+\AR{كم} you (plural)
+\AR{نا} us / our
+\AR{ي} me / my
\end{alltt}
\end{promptpart}

\begin{promptpart}{Surface-form normalization}
\begin{alltt}\footnotesize
Decliticization may result in malformed
surface forms.
In such cases, normalize tokens to their
naturally uncliticized form.
Alef-Lam
- \AR{للكتاب} → \AR{ل}+\AR{الكتاب}
(NOT: \AR{ل}+\AR{لكتاب})
Ta Marbuta
- \AR{مكتبتنا} → \AR{مكتبة} +\AR{نا}
(NOT: \AR{مكتبت} +\AR{نا})
Alef Maqsura
- \AR{مستشفاهم} → \AR{مستشفى} +\AR{هم}
(NOT: \AR{مستشفا} +\AR{هم})
Case-variant Hamza
- \AR{بهاوه} / \AR{بهاوه} / \AR{بهايه}
  → \AR{بهاء} +\AR{ه}
(NOT: \AR{بهاو} +\AR{ه} /
 \AR{بهاي} +\AR{ه})
Use "+" only to mark split clitics.
The produced token list must be used
exactly
as-is for dependency annotation.
\end{alltt}
\end{promptpart}

\begin{promptpart}{Query}
\begin{alltt}\footnotesize
Tokenize and dependency-parse this raw
sentence using CATiB.

RAW_TEXT:
<RAW TEXT>
\end{alltt}
\end{promptpart}
\newpage
\subsection{Cascaded Morphology-Guided Parsing Prompt}
\label{app:cascaded_prompt}
\noindent\textit{Composition note.}
The cascaded prompt combines blocks already shown in the preceding two
appendices. The CATiB task, raw-text tokenization and normalization rules,
output schema, dependency labels, and attachment rules are reused from
Appendix~\ref{app:parsing_prompts}. The names, meanings, and allowed values of
the 14 morphological fields are reused from
Appendix~\ref{app:mtag_prompt}. The rendered prompt included those shared
definitions in full; they are cross-referenced here to avoid duplicating them.

\begin{promptpart}{Auxiliary morphology block}
\begin{alltt}\footnotesize
You are also given a predicted word-level
morphosyntactic analysis for each
whitespace token in the raw sentence.
The predictions are auxiliary evidence
and may contain errors; the RAW TEXT is
authoritative.

Each morphology line contains the word
index, surface word, and only its
informative feature=value pairs. Values
meaning no information (0, na, u, empty,
or null) are omitted.
\end{alltt}

\noindent\textit{Inserted at this point:} the 14 morphological field
definitions and allowed values listed in
Appendix~\ref{app:mtag_prompt}.

\begin{alltt}\footnotesize
Use morphology to help decide clitic
splitting, token categories, and syntactic
attachments, but do not copy an analysis
when it conflicts with the raw form or
CATiB rules.
\end{alltt}
\end{promptpart}

\begin{promptpart}{Query}
\begin{alltt}\footnotesize
Now annotate the following Arabic
sentence using CATiB.
Sentence:
<RAW TEXT>

Predicted word-level morphology
(index, word, informative features):
<INDEX>  <WORD>  <FEATURE=VALUE>|...
\end{alltt}
\end{promptpart}

\newpage
\section{Licenses}
\label{app:license}

We list below the licenses of the data and tools used in this work, all of which are employed in accordance with their intended use.

\begin{itemize}[
  itemsep=1pt,
  parsep=1pt,
  topsep=2pt,
  partopsep=0pt
]

    \item \textbf{Penn Arabic Treebank (PATB)}~\citep{Maamouri:2004:penn}: LDC User Agreement. We use Parts~1, 2, and 3 (LDC2010T13, LDC2011T09, and LDC2010T08, respectively).\footnote{\url{https://catalog.ldc.upenn.edu/LDC2010T13}; \url{https://catalog.ldc.upenn.edu/LDC2011T09}; \url{https://catalog.ldc.upenn.edu/LDC2010T08}} Because our morphological and syntactic data is converted from licensed PATB data, we distribute the converted datasets in encrypted form generated using the Muddler tool developed by CAMeL Lab.\footnote{\url{https://github.com/CAMeL-Lab/muddler}} Reconstruction requires the corresponding licensed PATB source files.
    \item \textbf{CAMeL Treebank (CAMeLTB)}~\citep{habash-etal-2022-camel}: an open-source CAMeL Lab resource, accessed through its official distribution page.\footnote{\url{https://sites.google.com/nyu.edu/camel-treebank/resources}}
    \item \textbf{Llama~4~Scout}: Llama~4 Community License Agreement.\footnote{\url{https://huggingface.co/meta-llama/Llama-4-Scout-17B-16E-Instruct/blob/main/LICENSE}}
    \item \textbf{Qwen3-Next-80B-A3B-Thinking}: Apache License 2.0.\footnote{\url{https://huggingface.co/Qwen/Qwen3-Next-80B-A3B-Thinking/blob/main/LICENSE}}
    \item \textbf{Gemini~3~Flash} and \textbf{GPT-5.2}: accessed through provider APIs under the applicable Google Cloud and OpenAI service terms.\footnote{\url{https://cloud.google.com/terms/service-terms}; \url{https://openai.com/policies/services-agreement/}}
    \item \textbf{CAMeL Tools}~\citep{obeid-etal-2020-camel} and \textbf{CamelParser2.0}~\citep{elshabrawy-etal-2023-camelparser2}: MIT License.\footnote{\url{https://github.com/CAMeL-Lab/camel_tools/blob/master/LICENSE}; \url{https://github.com/CAMeL-Lab/camel_parser/blob/main/LICENSE}}

\end{itemize}

\clearpage 
\onecolumn

\section{Analysis}
\subsection{Tokenization Error Analysis}

\label{app:tokerr}
\begin{table*}[h!]
\centering
\small
\begin{tabular}{lccccc}
\toprule
\textbf{Error Type} & \multicolumn{2}{c}{\textbf{CAMeL~Tools}} & \multicolumn{2}{c}{\textbf{Gemini3}} & \textbf{Example} \\
\cmidrule{2-5}
 & Count & \% & Count & \% & \\
\midrule
Input Error & 35 & 9\% & 23 & 10\% &
\<الصحبة>  \textit{AlSHb{\TAMAR} } $\rightarrow$ \<الصحة>  \textit{AlSH{\TAMAR}}  \\

Gold Issues & 8 & 2\% & 13 & 6\% &
\<شعري>  \textit{{\SHIN}Ery} $\rightarrow$ \<شعر\ +ي>  \textit{{\SHIN}Er +y}  \\

Under Tokenization & 40 & 11\% & 8 & 4\% &
\<مدينة\ +ي>  \textit{mdyn{\TAMAR} +y} $\rightarrow$ \<مدينتي> \textit{mdynty} \\

Over Tokenization & 26 & 7\% & 7 & 3\% &
\<فيران> \textit{fyrAn} $\rightarrow$ \<ف+\ يران> \textit{f yrAn} \\

Substitution & 3 & 1\% & 4 & 2\% &
\<والدا\ +ي> \textit{wAldA +iy} $\rightarrow$ \<والد\ +ي> \textit{wAld +y} \\

Normalization & 56 & 15\% & 35 & 15\% &
\<آسف>  \textit{{\AMADDA}sf} $\rightarrow$ \<أسف> \textit{{\AHAMZAUP}sf} \\

Punctuation & 204 & 55\% & 10 & 4\% &
\<؟> $\rightarrow$ ? \\

Hallucination & 0 & 0\% & 128 & 56\% &
\<ب+\ مركبة> (b+\_markaba) $\rightarrow$ 
\<مركبة> \includegraphics[]{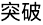}  \<ب+> \textit{b+\includegraphics[]{chinese-char.pdf} mrkb{\TAMAR}} \\

\midrule
\textbf{Total} & 372 & & 228 & & \\
\bottomrule
\end{tabular}
\caption{Distribution of tokenization errors for CAMeL~Tools (as used in CamelParser) and Gemini3 Best-ICL on a 500-sentence sample from the CAMeLTB Test split in the raw-text setting. Arabic examples are shown with Buckwalter-style transliteration.}
\label{tab:tok-errors}
\end{table*}

\subsection{Time \& Cost Analysis}
\label{app:time-money}
\begin{table*}[h!]
\centering
\small
\tabcolsep3pt
\begin{tabular}{llcccc}
\toprule
\textbf{Task} & \textbf{Model} & \multicolumn{2}{c}{\textbf{0}} & \multicolumn{2}{c}{\textbf{10}} \\
\cmidrule(lr){3-4} \cmidrule(lr){5-6}
 &  & \textbf{Time (min)} & \textbf{Cost (\$)} & \textbf{Time (min)} & \textbf{Cost (\$)} \\
\midrule

\textbf{Morphosyntactic Tagging}
& Llama4       & 2130.69 & 0.0   & 2084.87 & 0.0   \\
& Qwen3        & 1746.42 & 0.0   & 1294.90 & 0.0   \\
& GPT5.2       & 11.55   & 60.0  & 166.06  & 146.0 \\
& Gemini3      & 1595.76 & 64.0  & 1608.69 & 108.3 \\
& CAMeL Tools  & \textbf{1.17} & \textbf{0.0} & -- & -- \\
\midrule

{\textbf{Dependency Parsing (Gold Tokenization)}}
& Llama4       & 246.51  & 0.0   & 274.44  & 0.0   \\
& Qwen3        & 192.98  & 0.0   & 209.56  & 0.0   \\
& GPT5.2       & 22.53   & 22.0  & 23.41   & 33.0  \\
& Gemini3      & 1612.59 & 55.0  & 1070.54 & 41.0  \\
& CamelParser2.0 & \textbf{0.13} & \textbf{0.0} & -- & -- \\
\midrule

{\textbf{Dependency Parsing (Raw Text)}}
& Llama4       & 250.92  & 0.0   & 263.41  & 0.0   \\
& Qwen3        & 137.83  & 0.0   & 406.66  & 0.0   \\
& GPT5.2       & 32.64   & 34.0  & 33.93   & 41.0  \\
& Gemini3      & 2491.41 & 85.0  & 1668.20 & 61.8  \\
& CamelParser2.0 & \textbf{1.66} & \textbf{0.0} & -- & -- \\
\bottomrule
\end{tabular}
\caption{Cost and runtime comparison across tasks and models on Test sets under \textbf{0-shot} and \textbf{10-shot} settings. Supervised pipelines (CAMeL Tools and the original CamelParser2.0 pipeline) do not use in-context learning, so only a single runtime is reported.}

\end{table*}

\newpage 
\subsection{Data Per Shot}
\label{app:data-req}

\begin{table*}[h!]
\centering
\small 
\tabcolsep3.5pt
\begin{tabular}{llcccccccccc}
\toprule
\multirow{2}{*}{\textbf{Dataset}} & \textbf{} & \multicolumn{10}{c}{\textbf{Shots}} \\
\cmidrule(lr){3-12}
 &  & 1 & 2 & 3 & 4 & 5 & 6 & 7 & 8 & 9 & 10 \\
\midrule
PATB Dev  & Sents & 1,248 & 2,103 & 2,947 & 3,732 & 4,474 & 5,172 & 5,867 & 6,625 & 7,342 & 7,768 \\
PATB Dev  & Tokens & 47,932 & 80,514 & 113,086 & 143,733 & 172,664 & 200,312 & 227,903 & 256,349 & 282,103 & 296,309 \\
\midrule
PATB Test & Sents & 1,224 & 2,071 & 2,902 & 3,687 & 4,423 & 5,116 & 5,808 & 6,558 & 7,257 & 7,650 \\
PATB Test & Tokens & 46,801 & 78,853 & 110,776 & 140,919 & 169,451 & 196,693 & 223,925 & 252,001 & 277,566 & 290,559 \\
\midrule
CAMeLTB Dev  & Sents & 1,083 & 1,876 & 2,598 & 3,246 & 3,859 & 4,440 & 4,995 & 5,536 & 6,060 & 6,315 \\
CAMeLTB Dev  & Tokens & 14,982 & 25,781 & 35,567 & 44,450 & 52,768 & 60,615 & 68,052 & 75,265 & 82,082 & 85,292 \\
\midrule
CAMeLTB Test & Sents & 1,047 & 1,821 & 2,525 & 3,155 & 3,750 & 4,313 & 4,850 & 5,371 & 5,878 & 6,120 \\
CAMeLTB Test & Tokens & 14,402 & 24,916 & 34,473 & 43,134 & 51,258 & 58,904 & 66,141 & 73,155 & 79,791 & 82,930 \\
\bottomrule
\end{tabular}
\caption{Number of unique sentences and tokens required as the number of shots increases. }

\end{table*}

\subsection{Per Feature Accuracy}
\label{app:mtag_feature_accuracy}
\begin{table*}[h!]
\centering
\small
\tabcolsep1pt
\begin{tabular}{lccccccccccccccccc}
\toprule
\textbf{Model} & \textbf{Shots} & \multicolumn{1}{c}{\textbf{All Tags}} & \multicolumn{1}{c}{\textbf{Tag F\textsubscript{1}}} & \multicolumn{1}{c}{\textbf{POS}} & \multicolumn{1}{c}{\textbf{PER}} & \multicolumn{1}{c}{\textbf{GEN}} & \multicolumn{1}{c}{\textbf{NUM}} & \multicolumn{1}{c}{\textbf{ASP}} & \multicolumn{1}{c}{\textbf{MOD}} & \multicolumn{1}{c}{\textbf{VOX}} & \multicolumn{1}{c}{\textbf{STT}} & \multicolumn{1}{c}{\textbf{CAS}} & \multicolumn{1}{c}{\textbf{PRC0}} & \multicolumn{1}{c}{\textbf{PRC1}} & \multicolumn{1}{c}{\textbf{PRC2}} & \multicolumn{1}{c}{\textbf{PRC3}} & \multicolumn{1}{c}{\textbf{ENC0}} \\
\midrule
Llama4      & 0  & 7.7  & 78.9 & 71.2 & 92.5 & 77.2 & 79.8 & 95.6 & 92.1 & 95.1 & 64.6 & 49.4 & 62.2 & 75.8 & 80.2 & 86.3 & 83.0 \\
Llama4      & 10 & 57.4 & 90.1 & 82.7 & 96.2 & 85.9 & 88.3 & 96.4 & 96.2 & 96.4 & 80.7 & 72.4 & 87.3 & 94.1 & 93.8 & 96.5 & 94.0 \\
Qwen3       & 0  & 13.9 & 83.1 & 81.4 & 98.2 & 85.0 & 84.1 & 97.6 & 96.9 & 97.2 & 69.4 & 53.6 & 68.4 & 77.5 & 81.5 & 88.0 & 85.3 \\
Qwen3       & 10 & 76.5 & 97.0 & 95.0 & 99.4 & 95.4 & 96.0 & 99.4 & 99.1 & 99.0 & 92.9 & 87.0 & 97.7 & 99.3 & 99.4 & 99.9 & 98.7 \\
GPT5.2      & 0  & 39.7 & 89.9 & 91.1 & 99.2 & 89.4 & 90.0 & 99.7 & 94.3 & 99.5 & 84.1 & 81.9 & 73.3 & 87.6 & 89.4 & 89.7 & 89.2 \\
GPT5.2      & 10 & 86.2 & 98.4 & 96.7 & \textbf{99.8} & 96.5 & 96.5 & \textbf{99.8} & 99.6 & 99.6 & 97.2 & 93.2 & 99.1 & 99.6 & 99.7 & \textbf{100.0} & 99.7 \\
Gemini3     & 0  & 48.8 & 93.0 & 92.5 & 99.0 & 91.6 & 92.4 & \textbf{99.8} & 94.3 & 99.7 & 91.3 & 86.0 & 79.6 & 93.3 & 94.5 & 94.7 & 93.9 \\
Gemini3     & 10 & 86.6 & 98.5 & 96.9 & \textbf{99.8} & 96.4 & 95.9 & \textbf{99.8} & 99.7 & 99.7 & 97.7 & 94.6 & 99.4 & 99.6 & 99.7 & 99.9 & 99.7 \\
CAMeL Tools & -- & \textbf{96.3} & \textbf{99.6} & \textbf{98.8} & \textbf{99.8} & \textbf{99.8} & \textbf{99.8} & \textbf{99.8} & \textbf{99.8} & \textbf{99.8} & \textbf{99.1} & \textbf{97.7} & \textbf{99.8} & \textbf{99.9} & \textbf{99.9} & \textbf{100.0} & \textbf{99.9} \\
\bottomrule
\end{tabular}
\caption{Per-feature morphosyntactic tagging accuracy on the PATB Test split across models and prompting settings. The 10-shot rows use the Dev-selected highest-chrF++ retrieval configuration. GEN and NUM denote the form gender and form number features, respectively \cite{alkuhlani-habash-2011-corpus}.}
\label{tab:mtag_feature_accuracy}
\end{table*}


\subsection{Universal Dependencies Evaluation}
\label{app:nudar_ud}

\begin{table*}[h!]
\centering
\small
\setlength{\tabcolsep}{6pt}
\begin{tabular}{llcccccc}
\toprule
\textbf{Model} & \textbf{Setting} & \textbf{UPOS} & \textbf{All Tags} & \textbf{Tag F\textsubscript{1}} & \textbf{LS} & \textbf{UAS} & \textbf{LAS} \\
\midrule
Llama4  & Zero-shot & 87.3 & 43.2 & 77.7 & 49.7 & 45.3 & 28.7 \\
Llama4  & 10-shot   & 91.4 & 65.2 & 86.7 & 68.8 & 63.7 & 52.1 \\
\addlinespace[2pt]
Qwen3   & Zero-shot & 88.4 & 46.7 & 78.0 & 51.5 & 46.6 & 29.5 \\
Qwen3   & 10-shot   & 92.7 & 69.2 & 88.6 & 70.9 & 68.5 & 56.1 \\
\addlinespace[2pt]
GPT5.2  & Zero-shot & 91.5 & 56.4 & 84.5 & 61.3 & 74.4 & 53.2 \\
GPT5.2  & 10-shot   & 94.2 & 74.6 & 91.4 & 73.4 & 79.6 & 65.0 \\
\addlinespace[2pt]
Gemini3 & Zero-shot & 92.6 & 66.1 & 88.6 & 63.5 & 75.3 & 54.7 \\
Gemini3 & 10-shot   & 94.8 & 77.7 & 93.0 & 73.5 & 79.6 & 64.2 \\
\bottomrule
\end{tabular}
\caption{NUDAR/UD results on the Test split with gold tokenization. To examine whether the observed ICL trends extend beyond CATiB, we conduct an evaluation on the Arabic Universal Dependencies (NUDAR) Test split, a Universal Dependencies representation derived from PATB and CATiB annotations \citep{taji-etal-2017-universal}. Given gold-tokenized input, each LLM jointly predicts UPOS, 12 UD morphological features, dependency heads, and dependency relations for 1,963 sentences (74,125 syntactic tokens). All Tags requires exact agreement on UPOS and all 12 morphological features. We compare zero-shot prompting with 10-shot prompting, with the latter using the ten highest-chrF++ examples from the NUDAR Train split.}
\label{tab:nudar_ud}
\end{table*}


\newpage 
\subsection{LAS Breakdown by Genre\\ and Data Characteristics}
\label{app:genre-analysis}

\begin{table*}[h]
\centering
\small
\setlength{\tabcolsep}{5pt}
\begin{tabular}{l l l l c c c}
\toprule
\textbf{Genre} & \textbf{Variant} & \textbf{Period} & \textbf{Length} & \textbf{CamelParser2.0} & \textbf{CamelParser2.1} & \textbf{Gemini3} \\
\midrule
Odes      & CA  & 6th--12th  & Short & 77.3 & 77.6 & \textbf{83.3} \\
Quran     & CA  & 6th--12th  & Long  & 82.7 & 90.1 & \textbf{92.2} \\
Hadith    & CA  & 6th--12th  & Mid   & 92.4 & \textbf{92.7} & 90.6 \\
1001      & CA  & 6th--12th  & Mid   & 91.9 & \textbf{92.1} & 91.7 \\
Hayy      & CA  & 6th--12th  & Long  & 91.7 & 91.9 & \textbf{92.5} \\\midrule
OT        & MSA & 19th--20th & Long  & 87.4 & \textbf{91.7} & 88.5 \\
NT        & MSA & 19th--20th & Long  & 84.7 & \textbf{90.1} & 89.1 \\
Sara      & MSA & 19th--20th & Long  & 84.5 & \textbf{87.8} & 83.7 \\\midrule
ALC       & MSA & 21st       & Mid   & 87.5 & \textbf{87.7} & 84.7 \\
BTEC      & MSA & 21st       & Short & 84.9 & \textbf{86.9} & 86.8 \\
QALB      & MSA & 21st       & Mid   & \textbf{86.5} & 86.4 & 84.7 \\
WikiNews  & MSA & 21st       & Long  & \textbf{90.1} & \textbf{90.1} & 88.3 \\
ZAEBUC    & MSA & 21st       & Mid   & 91.8 & \textbf{92.3} & 92.0 \\
\specialrule{\lightrulewidth}{6pt}{4pt}
\multicolumn{7}{c}{\textbf{Macro Averages}}\\
\specialrule{\lightrulewidth}{4pt}{3pt}
 All Genres & & & & 87.2 & \textbf{89.0} & 88.3 \\
\midrule
CA        &     &        &       & 87.2 & 88.9 & \textbf{90.1} \\
MSA       &     &        &       & 87.2 & \textbf{89.1} & 87.2 \\
\midrule
6th--12th  &     &        &       & 87.2 & 88.9 & \textbf{90.1} \\
19th--20th &     &        &       & 85.5 & \textbf{89.9} & 87.1 \\
21st       &     &        &       & 88.2 & \textbf{88.7} & 87.3 \\
\midrule
Short (<10) &  &        &       & 81.1 & 82.3 & \textbf{85.1} \\
Mid (10--15)&  &        &       & 90.0 & \textbf{90.2} & 88.7 \\
Long (>15)  &  &        &       & 86.8 & \textbf{90.3} & 89.1 \\
\bottomrule
\end{tabular}
\caption{Labeled Attachment Score (LAS) by CAMeLTB genre for CamelParser2.0, CamelParser2.1, and Gemini3 under its best-performing ICL setting. The table also reports unweighted macro averages by variety, period, and input-length group, with the length-group thresholds shown in the final rows.}
\end{table*}

\end{document}

%% file: macros.tex
\newcommand{\hide}[1]{}

\newcommand{\AMADDA}{{\={A}}}
\newcommand{\AHAMZAUP}{{\^{A}}}

\newcommand{\TAMAR}{{$\hbar$}}

\newcommand{\SHIN}{{\v{s}}}

\newcommand{\AMAQ}{{\'{y}}}

%% file: Example.tex
{\begin{figure*}[th!]
\centering
  \includegraphics[width=0.8\textwidth]{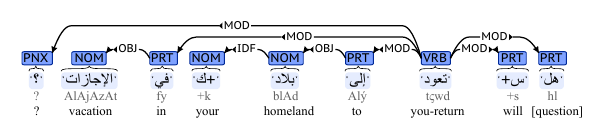}
\caption{Example \textbf{dependency parsing} output for “\textit{will you return to your homeland during vacation?}”. The model predicts a syntactic head (0 = root) and CATiB dependency relation for each token.}
\label{fig:parse_example}
\end{figure*}

\begin{table*}[th!]
\centering
\small
\tabcolsep4pt
\begin{tabular}{lccccccc}
\toprule
 \textbf{Words} & \<؟> & \<الإجازات> & \<في> & \<بلادك> & \<إلى> & \<ستعود> & \<هل> \\
\midrule
\textbf{Transliteration} & ? & Al\u{A}jAzAt & fy & blAdk & \u{A}l\'{y} & st$\varsigma$wd & hl \\
\textbf{Gloss} & ? & vacation & in & your-homeland & to & will-you-return & [question] \\
\midrule
pos  & punc & noun & prep & noun & prep & verb & part\_interrog \\
prc3 & na & 0 & 0 & 0 & 0 & 0 & 0 \\
prc2 & na & 0 & 0 & 0 & 0 & 0 & 0 \\
prc1 & na & 0 & 0 & 0 & 0 & 0 & 0 \\
prc0 & na & Al\_det & na & 0 & na & fut\_part & na \\
asp  & na & na & na & na & na & i & na \\
vox  & na & na & na & na & na & a & na \\
mod  & na & na & na & na & na & i & na \\
gen  & na & f & na & m & na & f & na \\
num  & na & p & na & p & na & s & na \\
stt  & na & d & na & c & na & na & na \\
cas  & na & g & na & g & na & na & na \\
per  & na & na & na & na & na & 3 & na \\
enc0 & na & 0 & 0 & poss\_2ms & 0 & 0 & 0 \\
\bottomrule
\end{tabular}
\caption{Example \textbf{morphosyntactic tagging} output for “\textit{will you return to your homeland during vacation?}”. The model predicts a 14-feature morphosyntactic analysis for each token.}\label{tab:morphtag}
\end{table*}}

%% file: data_stats.tex

\begin{table}[t]
\centering
\small
\tabcolsep8pt
\begin{tabular}{lcc}
\toprule
\textbf{Dataset} &\multicolumn{1}{c}{\textbf{Sentences}} & \multicolumn{1}{c}{\textbf{Words}} \\
\midrule
PATB Train & 15,000 & 477,512 \\
PATB Dev   & 1,986  & 63,137 \\
PATB Test  & 1,963  & 63,172 \\
\midrule
CAMeLTB Train & 9,397 & 133,459 \\
CAMeLTB Dev   & 2,022 & 27,464 \\
CAMeLTB Test  & 1,918 & 26,961 \\
\bottomrule
\end{tabular}
\caption{Dataset statistics for the morphosyntactic tagging and dependency parsing benchmarks.}
\label{tab:data_stats}
\end{table}
